\documentclass[runningheads]{llncs}

\usepackage{eccv}

\newcommand{\RNum}[1]{\uppercase\expandafter{\romannumeral #1\relax}}
\definecolor{darkgreen}{rgb}{0.167, 0.486, 0.347}
\definecolor{deemph}{gray}{0.55}
\usepackage{colortbl}
\newcommand{\grayrow}{\rowcolor[gray]{.9}}
\usepackage{pifont}
\usepackage{multirow, adjustbox}

\usepackage{eccvabbrv}

\usepackage{graphicx}
\usepackage{booktabs}

\usepackage[accsupp]{axessibility}  

\usepackage{hyperref}

\usepackage{orcidlink}

\usepackage{xspace}
\def\projmethod{\textit{SkillSpotter}\xspace}
\def\projmethodnormal{SkillSpotter\xspace}

\begin{document}

\title{\projmethodnormal: Pose-Aware Multi-View Skilled Action Detection and Grading in Ego-Exo Videos} 

\titlerunning{\projmethodnormal: Pose-Aware Multi-View Skilled Action Detection and Grading}


\author{Björn Braun\orcidlink{0000-0002-6027-0892} \and Christian Holz\orcidlink{0000-0001-9655-9519}
}

\authorrunning{B. Braun and C. Holz}

\institute{Department of Computer Science, ETH Zürich, Switzerland\\[1em]
\small\href{https://siplab.org/projects/SkillSpotter}{\color{magenta}{\texttt{https://siplab.org/projects/SkillSpotter}}}
}

\maketitle

\begin{abstract}

To enable personalized, real-time coaching using Augmented Reality glasses or fixed camera setups in domains such as sports, cooking, or music, a system must understand not just \emph{what} a person does, but \emph{how well} they execute an activity.
In an ego-exo video setting, this requires simultaneously detecting individual skilled actions and classifying each as correct or needing improvement, which Ego-Exo4D's  proficiency demonstration benchmark formalized.
We first adapt seven state-of-the-art temporal action detection architectures to this task, extend the evaluation protocol to disentangle detection from grading, and show that existing methods grade near-randomly.
We then introduce \projmethod, a pose-aware multi-view architecture that jointly detects and grades skilled actions through three task-specific modules:
(1)~adaptive temporal suppression to handle the varying density of skilled actions across diverse activities,
(2)~gated 3D body pose fusion to leverage body kinematics as a complementary signal to visual features, and
(3)~bidirectional cross-view attention to combine ego and exo views effectively.
\projmethod improves class-specific mAP from 12.40 to 21.82 (+76\%) and balanced accuracy from 55.99\% to 60.40\% over the best baseline.
\projmethod's modules transfer to other temporal action detection models with consistent gains and our method generalizes beyond Ego-Exo4D to HoloAssist.
\\[.2em]
Code: \url{https://github.com/eth-siplab/SkillSpotter}

\end{abstract}
\begin{figure}[t]
  \centering
  \includegraphics[width=\textwidth]{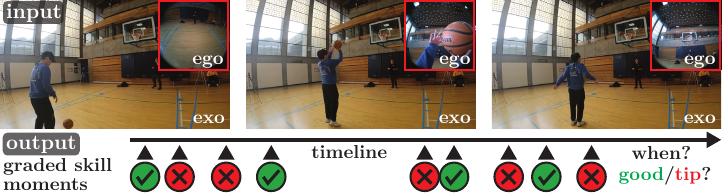}
    \caption{
      The \emph{proficiency demonstration} task requires detecting individual skilled actions from untrimmed synchronized ego-exo videos and classifying each action as \textit{good execution} (green) or \textit{tip for improvement} (red).
      \projmethod improves class-specific mean average precision from 12.40 to 21.82 (+76\%) and balanced accuracy from 55.99\% to 60.40\% over the best baseline.
}
  \label{fig:overview}
\end{figure}

\section{Introduction}
\label{sec:introduction}
Automatically assessing how well a person performs a skilled activity could enable personalized, real-time coaching through Augmented Reality (AR) glasses or fixed camera setups in domains such as sports, medical training, and music~\cite{wang2021survey,zhou2024comprehensive}.
This requires detecting \emph{when} skilled actions occur and determining \emph{how well} they are performed.
Temporal action detection (TAD) methods localize \emph{when} actions occur in untrimmed recordings by predicting temporal segments with start and end times~\cite{zhang2022actionformer,shi2023tridet,liu2022end,liu2024harnessing,liu2025opentad}, but do not assess execution quality.
Action quality assessment (AQA) methods score \emph{how well} a person performs an action from clips pre-cut around a single execution~\cite{zhou2024comprehensive,yu2021group,xu2024fineparser}.
However, neither of these tasks jointly detects and grades individual skill moments at the timestamp level in continuous recordings---a prerequisite for moment-by-moment feedback.

Ego-Exo4D~\cite{grauman2024ego} introduced this task in the \emph{proficiency demonstration} benchmark: detecting timestamps of skilled actions in ego-exo videos and additionally classifying each timestamp as good execution or needing improvement (see \cref{fig:overview}).
Only a minimal baseline exists for this task so far~\cite{grauman2024ego}.
We adapt seven state-of-the-art TAD architectures to establish a comprehensive baseline.
To separately evaluate detection and grading, we extend the evaluation protocol with class-agnostic mAP and balanced accuracy.
This reveals that current TAD architectures achieve low detection and grading quality near the random baseline.

In this paper, we introduce \projmethod, a pose-aware multi-view framework comprising three components that each address specific challenges of timestamp-level skill detection and grading.
\textit{Adaptive temporal suppression} addresses the large variation in action density across activities---from sub-second spacing in basketball to several seconds in music---by learning activity-specific suppression radii instead of using a fixed threshold.
\textit{Gated 3D body pose fusion} adds body kinematics to better assess execution quality.
\textit{Bidirectional cross-view attention} combines ego and exo views to prevent grading accuracy loss observed when naively concatenating these views.
We show that our modules transfer to other TAD architectures, and that \projmethod generalizes beyond Ego-Exo4D to HoloAssist.

\noindent We summarize our key contributions as follows:
\begin{enumerate}
    \item \textbf{\projmethod}, a pose-aware multi-view framework for timestamp-level skill detection and grading on the Ego-Exo4D proficiency benchmark, introducing three modules: adaptive temporal suppression, gated 3D body pose fusion, and bidirectional cross-view attention.
    Our method improves class-specific mAP from 12.40 to 21.82 (+76\%) and balanced accuracy from 55.99\% to 60.40\% over the strongest baseline, with consistent gains across all view settings and activities, generalizing beyond Ego-Exo4D to HoloAssist.
    \item \textbf{A comprehensive evaluation} of seven state-of-the-art TAD architectures on this task with an extended protocol that separately measures detection and quality grading, revealing that current architectures achieve low detection and near-random grading quality.
    Applying our proposed modules to these TAD architectures substantially improves their performance as well.
    \item \textbf{The first empirical inter-annotator agreement ceiling} for the Ego-Exo4D proficiency demonstration benchmark (BA: 64.6, Cohen's $\kappa$=0.29), based on the benchmark's multi-annotator data aligned with the evaluation protocol.
    The ceiling provides a calibration point for future work and shows that \projmethod reaches 94\% of it, suggesting that further BA gains are increasingly limited by label subjectivity rather than model capacity.
\end{enumerate}


\section{Related Work}
\label{sec:related_work}

\subsubsection{Skill Assessment in Ego-Exo Video.}
The Ego-Exo4D dataset~\cite{grauman2024ego} provides synchronized egocentric and exocentric video of skilled activities across eight scenarios with two proficiency benchmarks: \emph{demonstrator} proficiency (classifying a person's skill level) and \emph{demonstration} proficiency (timestamp-level detection and grading).
Recent works have focused on demonstrator proficiency estimation~\cite{bianchi2025skillformer, bianchi2025pats, wu2025skillsight} and language feedback generation~\cite{ashutosh2025expertaf, bianchi2025profvlm}.
Beyond visual features, egoPPG~\cite{braun2025egoppg} uses heart rate as a complementary cue for proficiency estimation.
The EgoExo-Fitness dataset~\cite{li2024egoexo} addresses a related problem in the fitness domain, providing segment-level proficiency annotations.
Unlike previous work that classifies demonstrator proficiency or generates language feedback, \projmethod jointly localizes and grades individual skilled actions by combining adaptive temporal suppression, 3D body pose, and ego-exo cross-view attention.

\subsubsection{Action Quality Assessment (AQA)} evaluates how well a person performs an action, typically by regressing a quality score from a pre-segmented clip~\cite{wang2021survey,zhou2024comprehensive,parmar2019and,tang2020uncertainty,yu2021group,xu2024fineparser}.
These methods assume known temporal boundaries, whereas Ego-Exo4D demonstration proficiency requires untrimmed timestamp localization and per-timestamp good-versus-tip classification.

\subsubsection{Temporal Action Detection and Action Spotting.}
Temporal action detection (TAD) localizes action segments in untrimmed video, evolving from two-stage proposal pipelines~\cite{lin2019bmn,lin2018bsn,zhao2021video,xu2020g} to one-stage anchor-free architectures~\cite{lin2021learning,zhang2022actionformer,shi2023tridet, yang2024dyfadet, tang2023temporalmaxer} and DETR-style set prediction~\cite{liu2022end} or hybrid Mamba--attention backbones~\cite{liu2024harnessing, chen2026video}, unified under the OpenTAD toolbox~\cite{liu2025opentad}.
These methods target segment localization with intersection-over-union (IoU)-based evaluation.
Action spotting also predicts single timestamps with tolerance-based matching~\cite{giancola2018soccernet,deliege2021soccernet}.
Recent methods use dense per-token classification with displacement heads~\cite{hong2022spotting,xarles2024t}.
The Ego-Exo4D demonstration proficiency benchmark shares this single-timestamp formulation but differs in three key aspects: it requires quality grading of each detected moment, operates on multi-view ego-exo input, and exhibits dense co-occurring events of different classes.

\subsubsection{Suppression and Fusion for Dense Events.}
Standard TAD pipelines apply Soft-NMS~\cite{bodla2017soft} to remove duplicate detections, with learned alternatives including density-aware thresholds~\cite{liu2019adaptive}, re-scoring networks~\cite{hosang2017learning}, and DETR-style set prediction~\cite{carion2020end} that removes NMS entirely.
None of these address timestamp-level, class-aware suppression for co-occurring feedback types.
For pose fusion, two-stream RGB--pose architectures are effective for action recognition~\cite{duan2022revisiting}, with 3D pose recoverable from ego-view head tracking~\cite{jiang2022avatarposer,jiang2024egoposerrobustreal} or multi-view triangulation~\cite{xu2022vitpose,iskakov2019learnable}; gated fusion~\cite{arevalo2017gated} and feature-wise modulation~\cite{perez2018film} offer lightweight integration.
Cross-attention across parallel view streams has been explored in representation learning~\cite{yan2022multiview,chen2021crossvit} but not for ego-exo skill assessment.
\section{Method}
\label{sec:method}

\subsection{Problem Formulation}
\label{sec:problem_formulation}
Given an untrimmed egocentric and exocentric video of a person performing a skilled activity, such as basketball, cooking, or music, the goal is to detect individual moments of correct or incorrect execution and classify each accordingly.
We formalize this following the \emph{demonstration proficiency estimation} task in Ego-Exo4D~\cite{grauman2024ego}, which jointly requires (i)~temporally localizing feedback timestamps and (ii)~classifying each as \emph{good execution} or \emph{tip for improvement}.

\subsubsection{Input.}
The input consists of synchronized egocentric and exocentric video features for a single task demonstration: $\mathbf{F}_{\text{ego}} \in \mathbb{R}^{C \times T}$ and $\{\mathbf{F}_{\text{exo}}^{m}\}_{m=1}^{M}$ with $\mathbf{F}_{\text{exo}}^{m} \in \mathbb{R}^{C \times T}$, where $C$ is the feature dimension and $T$ the number of temporal tokens.
Optionally, 3D body pose sequences $\mathbf{P} \in \mathbb{R}^{D_p \times T}$, where $D_p$ is the pose feature dimension, provide complementary skeletal information.

\subsubsection{Output.}
The model predicts a set of detections $\hat{\mathcal{D}} = \{(\hat{t}_k, \hat{y}_k, \hat{s}_k)\}_{k=1}^{K}$, where $\hat{t}_k \in \mathbb{R}^{+}$ is a predicted timestamp in seconds, $\hat{y}_k \in \{\texttt{good}, \texttt{tip}\}$ is the quality, and $\hat{s}_k \in [0, 1]$ is the confidence score.
Ground-truth annotations $\mathcal{G} = \{(t_j, y_j)\}_{j=1}^{J}$ consist of expert-annotated timestamps each labeled as a \emph{good execution} or a \emph{tip for improvement}.
Multiple annotations can share the same timestamp.

\subsubsection{Evaluation.}
\label{sec:evaluation}
For each radius $\delta \in \{0.25, 0.5, 1.0\}$\,s, the evaluator sorts predictions by confidence and greedily matches each prediction to the closest unmatched ground-truth timestamp if $|\hat{t}_k - t_j| \le \delta$.
The original benchmark reports class-specific mAP, which requires both correct localization and correct label assignment.
To decompose detection from classification, we \textbf{extend the evaluation protocol} with three complementary metrics:
(i)~\emph{class-agnostic mAP} (mAP\textsubscript{A}), which ignores the quality label and measures pure timestamp detection;
(ii)~\emph{class-specific mAP} (mAP\textsubscript{S}), equivalent to the original metric;
and (iii)~\emph{balanced accuracy} (BA) and \emph{macro-F1} (F1), computed on matched prediction--ground-truth pairs, which partially decouple classification quality from localization but still depend on matching coverage.

\subsection{Overall Architecture}
\label{sec:architecture}
\begin{figure}[tb]
  \centering
  \includegraphics[width=\textwidth]{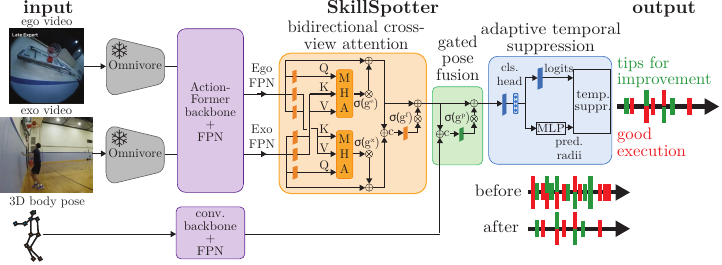}
    \caption{\projmethod architecture for skilled action detection and quality grading (Ego+Exos setting shown).
      \projmethod takes pre-extracted Omnivore features from ego and exo videos and estimated 3D body pose as input.
      Video features pass through an ActionFormer backbone and pose is encoded by a separate convolutional backbone.
      The output is a set of timestamped detections, each classified as good execution or tip for improvement.
      For single-view settings, the cross-view module is removed.
    }
  \label{fig:architecture}
\end{figure}

\projmethod extracts Omnivore~\cite{girdhar2022omnivore} video features from synchronized ego and exo videos, processes them with an ActionFormer~\cite{zhang2022actionformer} temporal backbone, and finally, predicts per-timestamp quality logits (see \cref{fig:architecture}).
We introduce three modules that each target a specific challenge of timestamp-level skill detection.
\emph{Bidirectional cross-view attention} (\cref{sec:crossview}) allows information exchange between ego and exo features.
\emph{Gated pose fusion} (\cref{sec:pose}) incorporates 3D body kinematics as a complementary signal for assessing execution quality.
\emph{Adaptive temporal suppression} (\cref{sec:suppression}) replaces fixed-radius non-maximum suppression (NMS) with learned, activity-specific suppression radii to address the large variation in skilled action density across activities.
Formally, the ActionFormer backbone processes the video features $\mathbf{F} \in \mathbb{R}^{C \times T}$ into a multi-scale temporal pyramid of $L{=}8$ levels $\{\mathbf{f}_l \in \mathbb{R}^{D \times T_l}\}_{l=0}^{L-1}$ with feature dimension $D{=}512$ and $T_l = T / 2^l$.
For single-view settings (Ego and Exos), cross-view attention is removed.

\subsubsection{Bidirectional Cross-View Attention.}
\label{sec:crossview}
Naively concatenating ego and exo features degrades grading performance for ActionFormer (see \cref{sec:experiments}).
We introduce bidirectional cross-view attention to enable information exchange between views.
The ego and exo feature streams are each processed through a shared-weight backbone into separate pyramids $\{\mathbf{f}_{\text{ego},l}\}$ and $\{\mathbf{f}_{\text{exo},l}\}$.
At each pyramid level $l$, both streams attend to each other via multi-head cross-attention ($N_h{=}4$ heads) with separate query, key, and value projections per direction:
\begin{align}
    \mathbf{f}_{\text{ego},l}' &= \mathbf{f}_{\text{ego},l} + \sigma(g_l^{\text{e}}) \cdot \text{MHA}(\mathbf{f}_{\text{ego},l},\, \mathbf{f}_{\text{exo},l}), \label{eq:cross_ego} \\
    \mathbf{f}_{\text{exo},l}' &= \mathbf{f}_{\text{exo},l} + \sigma(g_l^{\text{x}}) \cdot \text{MHA}(\mathbf{f}_{\text{exo},l},\, \mathbf{f}_{\text{ego},l}), \label{eq:cross_exo}
\end{align}
where $\text{MHA}(\mathbf{Q}, \mathbf{KV})$ denotes multi-head cross-attention with the first argument as queries and the second as keys and values.
Cross-view attention operates over the full temporal extent, unlike the backbone's windowed self-attention.
The enhanced features are merged into a unified representation:
\begin{equation}
    \mathbf{f}_l = \mathbf{f}_{\text{ego},l}' + \sigma(g_l^{f}) \cdot \text{Proj}([\mathbf{f}_{\text{ego},l}';\, \mathbf{f}_{\text{exo},l}']),
    \label{eq:cross_fusion}
\end{equation}
where the fusion gate $g_l^{f}$ is initialized to $-1.0$ and the residual path is from Ego, making the default model ego-centric.
Pose fusion (\cref{eq:pose_fusion}) is applied after cross-view fusion to integrate skeletal information into the unified representation.


\subsubsection{Gated Pose Fusion.}
\label{sec:pose}

Skill assessment is inherently tied to body mechanics.
We incorporate 3D body pose as a complementary signal via gated late fusion at each pyramid level.
Our default pose input $\mathbf{P} \in \mathbb{R}^{D_p \times T}$ with $D_p{=}118$ consists of 51 raw 3D keypoint coordinates (17 COCO joints $\times$ 3) and 67 kinematic features (joint angles, pairwise distances, and inter-frame velocities).
We train with the manually annotated GT Ego-Exo4D body pose~\cite{grauman2024ego} and evaluate with the pose that we predict ourselves, ensuring no ground-truth leakage at test time.
We report the GT-vs-predicted pose comparison in \cref{sec:supp_pose} (suppl.).
For the egocentric view, we predict body pose from the Aria camera trajectory using the official Ego-Exo4D baseline~\cite{grauman2024ego, castillo2023bodiffusion, jiang2022avatarposer,jiang2024egoposerrobustreal} (PA-MPJPE: 10.70\,cm).
For exocentric views, we detect the actor, estimate 2D keypoints with ViTPose~\cite{xu2022vitpose}, and recover 3D pose via calibrated multi-view triangulation with RANSAC-DLT (PA-MPJPE: 17.98\,cm).
We encode pose $\mathbf{P}$ with a convolutional backbone to 256 channels and map it to a temporal feature pyramid with output dimension $D{=}512$ per level.
At each level $l$, we fuse video and pose features via a gated residual:
\begin{equation}
    \mathbf{f}_l' = \mathbf{f}_l + \sigma(g_l^{p}) \cdot \text{Proj}([\mathbf{f}_l;\, \mathbf{p}_l])
    \label{eq:pose_fusion}
\end{equation}
where $\mathrm{Proj}:\mathbb{R}^{2D}\!\to\!\mathbb{R}^{D}$ is a $1{\times}1$ masked convolution followed by group normalization, ReLU, and dropout. $g_l^p$ is a learnable scalar gate initialized to $-2.0$.

\subsubsection{Adaptive Temporal Suppression.}
\label{sec:suppression}
\begin{figure}[t]
    \centering
    \includegraphics[width=0.5\linewidth]{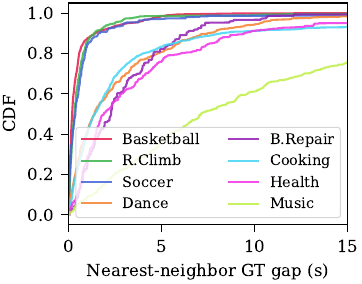}
    \caption{%
        Cumulative distribution of temporal distances between consecutive ground-truth events, shown per scenario.
        We see that the event density varies strongly across scenarios: in Basketball, over 80\% of events have a neighbor within 0.5\,s, whereas in Music fewer than 20\% do.
        This variation means no single fixed suppression radius can suit all scenarios, motivating our learned adaptive suppression.
    }
    \label{fig:density_cdf}
\end{figure}

Standard post-processing in TAD applies fixed-radius NMS to remove duplicate detections.
However, in skill assessment, events of different classes can co-occur at near or identical timestamps, and event density varies dramatically across scenarios: in Basketball, over 80\% of events have a neighbor within 0.5\,s, whereas in Music fewer than 20\% do (see \cref{fig:density_cdf}).
A single suppression threshold either erases valid co-occurring detections in dense scenarios or under-suppresses in sparse ones.

We replace fixed NMS with a learnable suppression mechanism that predicts per-detection suppression radii conditioned on the detection's features and scenario.
Unlike Adaptive NMS~\cite{liu2019adaptive}, our mechanism operates in the temporal domain with continuous exponential decay and class-aware constraints that allow co-occurring event types to coexist.
For each candidate detection $\mathbf{h}_k \in \mathbb{R}^{D}$ from the classification head, a two-layer MLP predicts a suppression radius:
\begin{equation}
    r_k = \text{Softplus}(\text{MLP}(\mathbf{h}_k)) + \mathbf{e}_{s},
    \label{eq:radius}
\end{equation}
where MLP : $\mathbb{R}^D \to \mathbb{R}^{128} \to \mathbb{R}^1$ with ReLU activation, and $\mathbf{e}_s \in \mathbb{R}$ is a per-scenario offset from a learnable embedding table ($S{=}8$ scenarios, indexed by the scenario label from the dataset metadata).

Given $K$ candidate detections sorted by confidence score $s_k$, each higher-scoring candidate $i$ suppresses lower-scoring candidate $j$ via exponential decay:
\begin{equation}
    w_{ij} = \exp\!\Big(\!-\frac{|t_i - t_j|}{r_i}\Big) \cdot \mathbb{1}[s_i > s_j] \cdot \mathbb{1}[y_i = y_j],
    \label{eq:suppression_weight}
\end{equation}
where the last indicator enforces \emph{class-aware} suppression: detections of different quality classes are not directly suppressed by this class-aware term, preserving co-occurring good/tip feedback at the same timestamp.
The adjusted score is $\tilde{s}_j = s_j \cdot \exp\!\big(\!-\sum_i w_{ij}\,/\,\tau\big)$, where $\tau$ is a learnable temperature initialized to 1.0.

To train the suppression module end-to-end, we introduce an auxiliary loss $\mathcal{L}_\text{sup}$ as the suppression module only operates at inference time and receives no gradient from the detection objective.
For each training video, we apply the suppressor to the classification scores to obtain adjusted scores $\tilde{s}_k$.
Gaussian targets centered on ground-truth timestamps define the supervision signal: $\hat{y}_k = \max_j \exp\!\big(\!-(t_k - t_j^*)^2 / 2\sigma_s^2\big)$ with $\sigma_s{=}1.0$ grid units.
The auxiliary loss is the binary cross-entropy between the adjusted scores and the Gaussian targets: $\mathcal{L}_{\text{sup}} = \text{BCE}(\text{logit}(\tilde{s}_k),\, \hat{y}_k)$.
The indicator $[\hat{s}_i > \hat{s}_j]$ and candidate selection act as fixed masks that determine suppression topology.
Gradients flow through the smooth exponential decay $\exp(-d_{ij} / r_i)$ to the radius predictor and temperature.

\subsubsection{Training Objective}
\label{sec:objective}
For classification, we use sigmoid focal loss~\cite{zhang2022actionformer}.
For temporal score alignment, we adopt the cumulative loss formulation of Kwak~et~al.~\cite{kwak2020detecting} used by the Ego-Exo4D baseline~\cite{grauman2024ego}:
\begin{equation}
  \mathcal{L}_{\text{reg}} = \frac{1}{N_{\text{pos}}}
    \sum_{c=1}^{N_c} \sum_{t}
    \Big(\mathrm{CDF}_{\text{pred}}^{c}(t)
       - \mathrm{CDF}_{\text{gt}}^{c}(t)\Big)^2,
  \label{eq:reg_loss}
\end{equation}
where $N_{\text{pos}}$ is the loss normalizer, $N_c$ is the number of classes, and $\mathrm{CDF}_{\text{pred}}^{c}(t)$ and $\mathrm{CDF}_{\text{gt}}^{c}(t)$ denote cumulative class-wise temporal score distributions.
The final loss combines all terms with $\lambda_{\text{reg}} = \lambda_{\text{sup}} = 1.0$:
\begin{equation}
  \mathcal{L} = \mathcal{L}_{\text{cls}}
    + \lambda_{\text{reg}}\,\mathcal{L}_{\text{reg}}
    + \lambda_{\text{sup}}\,\mathcal{L}_{\text{sup}}.
  \label{eq:total_loss}
\end{equation}

\subsection{Training}
\label{sec:implementation}
We train all models for 15 epochs (5 warmup epochs) with AdamW~\cite{loshchilov2017decoupled} (learning rate of $1 \times 10^{-3}$, weight decay of $0.05$) and a batch size of 8.
Training takes less than 15 minutes on one NVIDIA H200 GPU for all view settings.
Our full Ego+Exos model has 67.37M parameters and runs at 19.70 FPS on one H200 GPU.
See \cref{sec:supp_efficiency} (suppl.) for a full performance analysis.

\section{Experiments}
\label{sec:experiments}

\subsection{Dataset and Evaluation Protocol}
Following egoPPG~\cite{braun2025egoppg}, we use 90\% of the official training set of the \textit{proficiency demonstration} task for training, 10\% for validation, and the official validation set for testing.
As input, we use frozen Omnivore features ($C{=}1536$) to ensure fair comparison across all architectures and avoid overfitting the video backbone on the moderately sized training set (556 takes).
Takes without the required pre-extracted Omnivore features are excluded (see \cref{tab:supp_scenario_stats,tab:supp_scenario_stats_official} for statistics, suppl.).
The scenario sizes are highly imbalanced with 3--46 takes but classes near-balanced (49.5\% good).

We evaluate three view settings: \textit{Ego} (egocentric only), \textit{Exos} (four exocentric views), and \textit{Ego+Exos} (all five synchronized).
Following Ego-Exo4D~\cite{grauman2024ego}, we concatenate all feature streams for the Exos and Ego+Exos settings.
For our cross-view attention model (see~\cref{sec:crossview}), we keep ego and mean-pooled exo features as separate streams.
Single-view results without concatenation are in \cref{sec:supp_single_view} (suppl.).
All metrics are averaged over timestamp matching radii $\{0.25, 0.5, 1.0\}$\,s (see \cref{sec:evaluation}).
For fair comparison, all methods share the same Omnivore features and training protocol, and each baseline uses its default Soft NMS configuration as implemented in OpenTAD~\cite{liu2025opentad}.

\subsection{Evaluation Results}
\label{sec:main_results}
In \cref{tab:main}, we compare \projmethod to seven state-of-the-art TAD architectures~\cite{liu2022end, liu2024harnessing, zhang2022actionformer, shi2023tridet, chen2026video, yang2024dyfadet, tang2023temporalmaxer}. 
All re-implemented baselines substantially outperform the original benchmark result~\cite{grauman2024ego} (mAP$_\text{S}$: 3.27 vs.\ 7.63--12.40) but remain below 12.40 class-specific mAP, with detection limited to 8.65--17.11 mAP$_\text{A}$ and quality classification within 6 percentage points of the random baseline (BA 49.51--55.99\% vs.\ 50.9\% random).
The naive baselines follow the Ego-Exo4D protocol~\cite{grauman2024ego} (see \cref{tab:main}).
F1 tracks BA closely throughout all experiments; we focus on BA hereafter.
To reduce the chance that weak baseline performance is caused by under-tuning, we sweep TadTR learning rates/query counts (see \cref{sec:supp_tadtr}, suppl.) and compare single-view vs.\ concatenated inputs for four of the implemented baselines (see \cref{sec:supp_single_view}, suppl.).
These sweeps do not close the gap to \projmethod.
We additionally evaluate if AQA heads are better suited for this task than our classification head by substituting three AQA heads into our pipeline (see \cref{sec:suppl_aqa_heads}, suppl.). 
Our default head remains strongest in grading quality, while detection stays comparable across heads.

\projmethod improves class-specific mAP from 12.40 to 21.82 (+76\%) and balanced accuracy from 55.99\% to 60.40\% on Ego (see \cref{tab:main}).
These gains generalize across all view settings: Ego (21.82 mAP$_\text{S}$, 60.40 BA), Exos (21.12, 60.59), and Ego+Exos (21.34, 60.39), outperforming all baselines by a large margin.
Notably, some baselines degrade under naive Ego+Exos concatenation (e.g., ActionFormer BA 55.99$\to$50.34) compared to only using a single view as input.
We analyze this instability in \cref{sec:ablation}.

\begin{table}
    \centering
    \caption{Timestamp-level skill assessment on Ego-Exo4D demonstration proficiency.
    We report class-specific mean average precision (mAP\textsubscript{S}), class-agnostic mAP (mAP\textsubscript{A}), balanced accuracy (BA), and macro-F1 across three view settings.
    All metrics are averaged over matching radii $\{0.25, 0.5, 1.0\}$\,s.
    Baselines use Soft NMS.
    Our method uses learned adaptive suppression.
    $^\dagger$~Original benchmark result~\cite{grauman2024ego}, for which only mAP\textsubscript{S} was reported.
    $^*$~Uses cross-view attention for Ego+Exos.}
    \label{tab:main}
    \resizebox{\columnwidth}{!}{
    \begin{tabular}{@{}l cccc cccc cccc@{}}
        \toprule[1.5pt]
        & \multicolumn{4}{c}{\textbf{Ego}} & \multicolumn{4}{c}{\textbf{Exos}} & \multicolumn{4}{c}{\textbf{Ego+Exos}} \\
        \cmidrule(lr){2-5} \cmidrule(lr){6-9} \cmidrule(lr){10-13}
        \textbf{Model} & mAP$_\text{S}$ & mAP$_\text{A}$ & BA & F1 & mAP$_\text{S}$ & mAP$_\text{A}$ & BA & F1 & mAP$_\text{S}$ & mAP$_\text{A}$ & BA & F1 \\
        \midrule
        \midrule
        Random & 0.73 & 1.49 & 50.90 & 50.03 & 0.70 & 1.47 & 50.44 & 49.59 & 0.70 & 1.46 & 50.15 & 49.39 \\
        Uniform tips & 0.71 & 1.49 & 50.00 & 27.15 & 0.71 & 1.47 & 50.00 & 27.15 & 0.72 & 1.46 & 50.00 & 27.15 \\
        Uniform good & 0.70 & 1.52 & 50.00 & 38.55 & 0.68 & 1.47 & 50.00 & 38.55 & 0.67 & 1.46 & 50.00 & 38.55 \\
        \midrule
        Baseline$^\dagger$ & 3.27 & -- & -- & -- & 3.84 & -- & -- & -- & 3.57 & -- & -- & -- \\
        VideoMambaSuite~\cite{chen2026video} & 7.63  & 8.65  & 49.51 & 49.35 & 7.06 & 8.17 & 46.03 & 43.88 & 3.69 & 3.91 & 52.70 & 51.96\\
        TadTR~\cite{liu2022end} & 7.79 & 10.79 & 49.68 & 33.62 & 6.37 & 10.48 & 52.31 & 34.63 & 4.12 & 6.74 & 52.81 & 48.23 \\
        DyFADet~\cite{yang2024dyfadet} & 10.09 & 12.53 & 48.89 & 46.75 & 3.57 & 4.48 & 49.52 & 47.72 & 3.18 & 5.64 & 47.63 & 35.45\\
        TriDet~\cite{shi2023tridet} & 10.35 & 14.79 & 49.17 & 40.10 & 8.99 & 12.16 & 50.06 & 36.78 & 8.23 & 11.78 & 48.92 & 48.77 \\
        CausalTAD~\cite{liu2024harnessing} & 11.42 & 16.07 & 52.86 & 52.07 & 11.82 & 14.05 & 50.78 & 50.41 & 13.16 & 17.21 & 54.98 & 54.95 \\
        TemporalMaxer~\cite{tang2023temporalmaxer} & 12.34 & 16.69 & 54.34 & 53.90 & 10.38 & 15.17 & 52.18 & 52.16 & 11.27 & 15.75 & 50.09 & 49.58 \\
        ActionFormer~\cite{zhang2022actionformer} & 12.40 & 17.11 & 55.99 & 55.91 & 13.18 & 18.25 & 55.03 & 55.03 & 13.82 & 17.85 & 50.34 & 50.17 \\
        \midrule
        \grayrow\textbf{\projmethod}$^*$ & \textbf{21.82} & \textbf{27.89} & \textbf{60.40} & \textbf{60.02} & \textbf{21.12} & \textbf{27.47} & \textbf{60.59} & \textbf{60.55} & \textbf{21.34} & \textbf{28.01} & \textbf{60.39} & \textbf{59.37} \\
        \midrule
        $\Delta$ best baseline & \textcolor{darkgreen}{\texttt{+}\textbf{9.42}} & \textcolor{darkgreen}{\texttt{+}\textbf{10.78}} & \textcolor{darkgreen}{\texttt{+}\textbf{4.41}} & \textcolor{darkgreen}{\texttt{+}\textbf{4.11}} & \textcolor{darkgreen}{\texttt{+}\textbf{7.94}} & \textcolor{darkgreen}{\texttt{+}\textbf{9.22}} & \textcolor{darkgreen}{\texttt{+}\textbf{5.56}} & \textcolor{darkgreen}{\texttt{+}\textbf{5.52}} & \textcolor{darkgreen}{\texttt{+}\textbf{7.52}} & \textcolor{darkgreen}{\texttt{+}\textbf{10.16}} & \textcolor{darkgreen}{\texttt{+}\textbf{5.41}} & \textcolor{darkgreen}{\texttt{+}\textbf{4.42}} \\
        \bottomrule[1.5pt]
    \end{tabular}
    }
\end{table}

\subsection{Ablation Study}
\label{sec:ablation}

\subsubsection{Suppression Strategy.}
Removing Soft NMS entirely (No NMS) from the ActionFormer baseline~\cite{zhang2022actionformer} increases mAP$_\text{A}$ substantially (Ego: 16.85$\to$22.50) but mAP$_\text{S}$ only modestly (12.12$\to$13.96), while BA decreases slightly (55.34$\to$54.81).
This means that the additional detections recovered without suppression are poorly classified, indicating a trade-off between detection and grading precision.

Our introduced temporal adaptive suppression outperforms both fixed strategies (No NMS and Soft NMS) by a large margin: mAP$_\text{S}$ rises from 13.96 to 18.82, mAP$_\text{A}$ from 22.50 to 25.87, and BA from 54.81 to 59.74---jointly improving detection and classification.
Class-aware decay restricts suppression to detections of the same class (see \cref{eq:suppression_weight}), allowing temporally overlapping good and tip events to coexist.
We sweep Soft NMS $\sigma$ in \cref{sec:supp_soft_nms_sigma_sweep} (suppl.), showing that no single fixed $\sigma$ jointly optimizes detection and classification in our setting, and that the best fixed setting remains below adaptive suppression.

\begin{figure}[t]
    \centering
    \includegraphics[width=\linewidth]{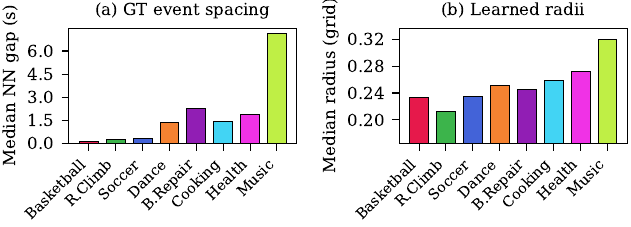}
    \caption{%
        (a) On Ego-Exo4D, the temporal distance between ground-truth events varies strongly across scenarios.
        (b) \projmethod learns to match this variation through our proposed adaptive temporal suppression module, which learns to predict scenario-specific suppression radii.
        Dense scenarios, such as Basketball, receive smaller radii, sparse ones, such as Music, larger radii (Spearman $\rho = 0.83$, $p = 0.010$).
    }
    \label{fig:density_radii}
\end{figure}

\cref{fig:density_radii} shows the ground-truth event density (left) and the median learned suppression radii (right) per scenario of our temporal suppression module.
The median learned radius per scenario correlates with the ground-truth event spacing (Spearman $\rho = 0.83$, $p = 0.010$): without any explicit density supervision, the model learns smaller radii for densely labeled scenarios such as Basketball (median nearest-neighbor gap 0.05\,s, radius 0.23 grid units) and larger radii for sparser ones such as Music (7.05\,s, 0.32 grid units).
Per-scenario radius distributions are provided in \cref{sec:supp_radius_boxplot} (suppl.).

\subsubsection{Pose Fusion.}
Adding 3D body pose further improves both detection and classification.
On Ego, mAP$_\text{S}$ increases from 18.82 to 21.82 (+3.00), mAP$_\text{A}$ from 25.87 to 27.89 (+2.02), and BA from 59.74 to 60.40.
On Exos, the BA gain is even larger (58.29$\to$60.59, +2.30), suggesting that pose is particularly complementary to exo-view appearance features.
We also evaluate whether predicted pose compared to ground truth during inference affects performance and find only a marginal decrease when using our predicted pose from the ego and exo views (see \cref{sec:supp_pose}, suppl.).
Adding hand-pose (using Ego-Exo4D's POTTER baseline) to our gated mechanism gives mAP$_\text{S}$ $21.82{\rightarrow}20.92$, BA $60.40{\rightarrow}59.56$ on Ego. 
We therefore did not include hand pose further.

\subsubsection{Cross-View Attention.}
While adaptive suppression and pose fusion improve all metrics in the single-view settings, the Ego+Exos setting exhibits a classification collapse: BA drops from 51.79 to 45.77 with adaptive suppression and pose fusion.

Cross-view attention addresses this by maintaining view-specific feature streams with explicit bidirectional information exchange before gated fusion, rather than relying on the backbone to separate concatenated ego-exo representations.
BA recovers to 60.39 (+14.62 over the concatenated), which is close to the Ego/Exos single-stream results (60.40/60.59 in \cref{tab:ablation}) and higher than the Ego+Exos per-camera single-view variant (55.93 in \cref{tab:supp_views}, suppl.).
We observe a slight mAP$_\text{S}$ decrease (21.99$\to$21.34) relative to the pose-only concatenated row, indicating a minor detection-grading tradeoff.
These results are consistent with the benefit of view-specific processing with explicit bidirectional exchange before fusion.

\begin{table}
\centering
\caption{Component ablation of our model. All rows use the ActionFormer backbone. Each row progressively adds one component. $^*$~Cross-view attention applies to Ego+Exos only; Ego and Exos results are unchanged from the row above.}
\label{tab:ablation}
\resizebox{\columnwidth}{!}{
\begin{tabular}{@{}l cccc cccc cccc@{}}
\toprule[1.5pt]
& \multicolumn{4}{c}{\textbf{Ego}} & \multicolumn{4}{c}{\textbf{Exos}} & \multicolumn{4}{c}{\textbf{Ego+Exos}} \\
\cmidrule(lr){2-5} \cmidrule(lr){6-9} \cmidrule(lr){10-13}
\textbf{Configuration} & mAP$_\text{S}$ & mAP$_\text{A}$ & BA & F1 & mAP$_\text{S}$ & mAP$_\text{A}$ & BA & F1 & mAP$_\text{S}$ & mAP$_\text{A}$ & BA & F1 \\
\midrule
\midrule
ActionFormer (Soft NMS) & 12.12 & 16.85 & 55.34 & 55.24 & 13.15 & 18.13 & 54.75 & 54.75 & 13.71 & 17.86 & 50.24 & 50.01 \\
ActionFormer (No NMS) & 13.96 & 22.50 & 54.81 & 54.26 & 15.42 & 24.47 & 54.85 & 54.65 & 17.30 & 24.12 & 51.79 & 51.40 \\
\quad + Adaptive Supp. & 18.82 & 25.87 & 59.74 & 59.47 & 21.03 & 27.14 & 58.29 & 57.65 & 21.18 & 27.94 & 43.06 & 39.39 \\
\quad + Pose Fusion & \textbf{21.82} & \textbf{27.89} & \textbf{60.40} & \textbf{60.02} & \textbf{21.12} & \textbf{27.47} & \textbf{60.59} & \textbf{60.55} & \textbf{21.99} & \textbf{28.04} & 45.77 & 43.57 \\
\quad + Cross-View Attn$^*$ (Ours) & -- & -- & -- & -- & -- & -- & -- & -- & 21.34 & 28.01 & \textbf{60.39} & \textbf{59.37} \\
\bottomrule[1.5pt]
\end{tabular}
}
\end{table}

\subsubsection{Cross-Backbone Evaluation.}
\label{sec:cross_backbone}
We also apply adaptive suppression and pose fusion to all implement backbones, except TadTR~\cite{liu2022end} (as it lacks a suppression stage), to test whether our modules transfer beyond ActionFormer (see \cref{tab:cross_backbone} and \cref{sec:suppl_cross_backbone2}, suppl.).
Adaptive suppression and pose fusion improve detection (mAP$_\text{S}$, mAP$_\text{A}$) across all backbones except VideoMambaSuite, for which only Adaptive Suppression improves performance.
Pose fusion's benefit to grading (BA, F1) is less consistent: it improves BA across all view settings only for TriDet and CausalTAD, while the other three each show a BA regression in at least one view setting.

\begin{table}
    \centering
    \caption{\projmethod's adaptive suppression and pose fusion transfer to five further TAD backbones on Ego-Exo4D demonstration proficiency.
    TriDet~\cite{shi2023tridet}, CausalTAD~\cite{liu2024harnessing}, VideoMambaSuite~\cite{chen2026video}, and TemporalMaxer~\cite{tang2023temporalmaxer} improve substantially across all view settings; DyFADet~\cite{yang2024dyfadet} improves detection throughout but suffers a grading regression on Exos ($^\dagger$).
    We exclude TadTR~\cite{liu2022end} because its DETR-style set prediction removes the suppression stage.
    }
    \label{tab:cross_backbone}
    \vspace{-2mm}
    \resizebox{\columnwidth}{!}{
    \begin{tabular}{@{}ll cccc cccc cccc@{}}
        \toprule[1.5pt]
        & & \multicolumn{4}{c}{\textbf{Ego}} & \multicolumn{4}{c}{\textbf{Exos}} & \multicolumn{4}{c}{\textbf{Ego+Exos}} \\
        \cmidrule(lr){3-6} \cmidrule(lr){7-10} \cmidrule(lr){11-14}
        \textbf{Backbone} & \textbf{Configuration} & mAP$_\text{S}$ & mAP$_\text{A}$ & BA & F1 & mAP$_\text{S}$ & mAP$_\text{A}$ & BA & F1 & mAP$_\text{S}$ & mAP$_\text{A}$ & BA & F1 \\
        \midrule
        \midrule
        \multirow{3}{*}{TriDet~\cite{shi2023tridet}} 
        & Soft NMS & 10.35 & 14.79 & 49.17 & 40.10 & 8.99 & 12.16 & 50.06 & 36.78 & 8.23 & 11.78 & 48.92 & 48.77 \\
        & No NMS & 11.80 & 17.80 & 52.03 & 50.56 & 11.47 & 15.45 & 48.53 & 48.49 & 9.46 & 14.71 & 48.83 & 48.25 \\
        & + Our modules & \textbf{20.47} & \textbf{26.91} & \textbf{59.98} & \textbf{58.77} & \textbf{19.21} & \textbf{24.32} & \textbf{60.56} & \textbf{59.93} & \textbf{19.33} & \textbf{25.56} & \textbf{57.85} & \textbf{57.84} \\
        \midrule
        \multirow{3}{*}{CausalTAD~\cite{liu2024harnessing}} 
        & Soft NMS & 11.42 & 16.07 & 52.86 & 52.07 & 11.82 & 14.05 & 50.78 & 50.41 & 13.16 & 17.21 & 54.98 & 54.95 \\
        & No NMS & 12.77 & 22.71 & 51.96 & 50.20 & 12.57 & 18.91 & 53.16 & 52.87 & 12.69 & 21.45 & 54.63 & 54.47 \\
        & + Our modules & \textbf{21.36} & \textbf{27.46} & \textbf{59.34} & \textbf{59.23} & \textbf{20.76} & \textbf{25.81} & \textbf{60.53} & \textbf{60.36} & \textbf{20.02} & \textbf{26.83} & \textbf{60.01} & \textbf{59.51} \\
        \midrule
        \multirow{3}{*}{VideoMambaSuite~\cite{chen2026video}} 
        & Soft NMS & 7.63 & 8.65 & 49.51 & 49.35 & 7.06 & 8.17 & 46.03 & 43.88 & 3.69 & 3.91 & 52.70 & 51.96 \\
        & No NMS & 13.60 & 16.22 & 53.55 & 53.23 & 11.44 & 15.70 & 47.79 & 40.48 & 6.91 & 9.22 & 52.02 & 47.96 \\
        & + Our modules & \textbf{21.39} & \textbf{27.11} & \textbf{55.01} & \textbf{53.50} & \textbf{18.66} & \textbf{26.33} & \textbf{54.46} & \textbf{54.06} & \textbf{19.91} & \textbf{26.03} & \textbf{56.92} & \textbf{56.90} \\
        \midrule
        \multirow{3}{*}{TemporalMaxer~\cite{tang2023temporalmaxer}} 
        & Soft NMS & 12.34 & 16.69 & 54.34 & 53.90 & 10.38 & 15.17 & 52.18 & 52.16 & 11.27 & 15.75 & 50.09 & 49.58 \\
        & No NMS & 14.76 & 22.83 & 53.89 & 53.12 & 11.56 & 20.53 & 52.57 & 52.55 & 13.20 & 22.42 & 52.07 & 50.46 \\
        & + Our modules & \textbf{20.15} & \textbf{27.85} & \textbf{58.66} & \textbf{57.48} & \textbf{20.65} & \textbf{27.01} & \textbf{59.66} & \textbf{59.66} & \textbf{20.55} & \textbf{27.38} & \textbf{59.42} & \textbf{59.34} \\
        \midrule
        \multirow{3}{*}{DyFADet~\cite{yang2024dyfadet}} 
        & Soft NMS & 10.09 & 12.53 & 48.89 & 46.75 & 3.57 & 4.48 & \textbf{49.52} & \textbf{47.72} & 3.18 & 5.64 & 47.63 & 35.45 \\
        & No NMS & 12.74 & 16.26 & 50.14 & 46.48 & 3.68 & 4.69 & 47.23 & 45.00 & 3.69 & 5.57 & 47.54 & 40.41 \\
        & + Our modules & \textbf{21.56} & \textbf{28.68} & \textbf{56.77} & \textbf{56.51} & \textbf{20.16} & \textbf{25.22} & 41.83$^\dagger$ & 36.71$^\dagger$ & \textbf{20.94} & \textbf{27.25} & \textbf{58.86} & \textbf{58.49} \\
        \bottomrule[1.5pt]
    \end{tabular}
    }
\end{table}

\subsection{Analysis}
\label{sec:analysis}
\subsubsection{Per-Scenario Breakdown.}
\cref{tab:analysis} (top) reports Ego results per scenario for our method and ActionFormer.
We compare against ActionFormer because it provides the strongest baseline trade-off in the Ego setting.
Per-scenario trends are consistent across view settings.
Our method improves mAP$_\text{S}$ and mAP$_\text{A}$ across all eight scenarios, with the largest mAP$_\text{S}$ gains on Basketball (+17.89) and Rock Climbing (+10.02).
BA improves by up to +9.68 (Dance), with the exception of Cooking, Health, and Soccer where ActionFormer retains a higher BA.
The largest gains appear in Basketball, Rock Climbing, and Dance, which are also among the denser scenarios (see \cref{fig:density_radii}).
We also observe that scenarios with coarse full-body motion benefit more than scenarios dominated by subtle manipulation cues.
This pattern is consistent with adaptive suppression helping dense-event regimes and with the current feature set (appearance + body pose) capturing gross kinematics better than fine hand-object interactions.
Cooking and Music remain challenging for detection (mAP$_\text{S}$: 3.08 and 1.57 for our method), indicating that current appearance-plus-body-pose cues still miss part of the quality signal.
In particular, fine-grained hand-object interactions likely require stronger hand-level or object-state representations than we currently use.
Soccer (BA: 49.48) and Cooking (F1: 43.22) remain difficult grading regimes.
We interpret Soccer cautiously because the evaluation split contains only three videos (see \cref{tab:supp_scenario_stats}, suppl.).
We additionally report detection recall~(R), defined as the fraction of ground-truth events matched by any prediction within the matching radius.
Our method consistently achieves higher recall than ActionFormer across all scenarios (e.g., 72.89\% vs.\ 49.69\% on Basketball), confirming that the BA and F1 improvements are not artifacts of selective matching but reflect genuinely better coverage of annotated events.

\subsubsection{Per-Radius Breakdown.}
\cref{tab:analysis} (bottom) reports results at individual matching radii for \projmethod and ActionFormer on Ego.
\projmethod improves mAP$_\text{S}$ over ActionFormer at every radius, with the largest relative gain at the strictest threshold (13.49 vs.\ 4.21 at 0.25\,s).
For both methods, BA decreases at looser radii while mAP$_\text{S}$ increases: stricter matching retains temporally precise detections, while looser matching admits predictions farther from annotated timestamps.
Detection recall (R) confirms that this BA drop is not an artifact of selective matching: at 0.25\,s only 38.92\% of GT events are matched, rising to 63.09\% at 1.0\,s, so the lower BA at loose radii reflects genuinely harder cases entering the evaluation pool.
\projmethod consistently matches more GT events than ActionFormer at every radius (e.g., 38.92\% vs.\ 21.28\% at 0.25\,s), indicating that its mAP and BA gains are not due to cherry-picking easy timestamps.
The mAP$_\text{A}$--mAP$_\text{S}$ gap widens at looser radii for both methods (e.g., Ours: 3.57 at 0.25\,s vs.\ 8.38 at 1.0\,s).

\begin{table}[t]
\centering
\caption{Per-scenario and per-radius breakdown of \projmethod vs.\ ActionFormer (Ego setting).
    \projmethod improves detection (mAP\textsubscript{S}, mAP\textsubscript{A}) and recall (R) across all eight scenarios, with the largest gains in dense, full-body scenarios such as Basketball and Rock Climbing.
    Per-radius results show that \projmethod is especially effective at the strictest matching threshold (0.25\,s).
    }
\label{tab:analysis}
\resizebox{\columnwidth}{!}{
\begin{tabular}{@{}l ccccc ccccc@{}}
\toprule[1.5pt]
& \multicolumn{5}{c}{\textbf{\projmethod (Ours)}} & \multicolumn{5}{c}{\textbf{ActionFormer}} \\
\cmidrule(lr){2-6} \cmidrule(lr){7-11}
& mAP$_\text{S}$ & mAP$_\text{A}$ & BA & F1 & R & mAP$_\text{S}$ & mAP$_\text{A}$ & BA & F1 & R \\
\midrule
\midrule
\multicolumn{11}{@{}l}{\textit{Per Scenario}} \\
\midrule
Basketball & \textbf{40.76} & \textbf{49.02} & \textbf{59.78} & \textbf{59.49} & \textbf{72.89} & 22.87 & 29.60 & 52.69 & 52.66 & 49.69 \\
Bike Repair & \textbf{7.65} & \textbf{10.98} & \textbf{63.09} & \textbf{62.13} & \textbf{51.13} & 3.01 & 4.45 & 53.56 & 52.10 & 33.68 \\
Cooking & \textbf{3.08} & \textbf{6.75} & 51.64 & 43.22 & \textbf{27.32} & 1.94 & 4.34 & \textbf{52.02} & \textbf{48.02} & 21.37 \\
Dance & \textbf{8.02} & \textbf{11.58} & \textbf{61.27} & \textbf{59.69} & \textbf{52.75} & 5.79 & 9.58 & 51.59 & 46.99 & 37.92 \\
Health & \textbf{2.67} & \textbf{3.96} & 56.24 & 56.18 & \textbf{20.05} & 1.70 & 2.75 & \textbf{59.97} & \textbf{57.91} & 18.65 \\
Music & \textbf{1.57} & \textbf{1.91} & \textbf{58.94} & \textbf{58.90} & \textbf{23.03} & 0.41 & 0.89 & 58.19 & 54.70 & 15.74 \\
Rock Climbing & \textbf{28.87} & \textbf{42.47} & \textbf{58.84} & \textbf{55.94} & \textbf{73.86} & 18.85 & 26.79 & 52.90 & 51.59 & 53.05 \\
Soccer & \textbf{12.99} & \textbf{24.54} & 49.48 & 36.57 & \textbf{37.02} & 7.60 & 12.20 & \textbf{58.37} & \textbf{57.76} & 27.11 \\
\midrule
\multicolumn{11}{@{}l}{\textit{Per Radius}} \\
\midrule
0.25s & \textbf{13.49} & \textbf{17.06} & \textbf{64.75} & \textbf{64.89} & \textbf{38.92} & 4.21 & 6.89 & 54.70 & 54.67 & 21.28 \\
0.5s & \textbf{21.87} & \textbf{28.15} & \textbf{59.79} & \textbf{59.28} & \textbf{52.21} & 10.21 & 14.62 & 55.76 & 55.69 & 34.90 \\
1.0s & \textbf{30.09} & \textbf{38.47} & \textbf{56.65} & \textbf{55.89} & \textbf{63.09} & 21.93 & 29.05 & 55.55 & 55.37 & 53.25 \\
\midrule
Average & \textbf{21.82} & \textbf{27.89} & \textbf{60.40} & \textbf{60.02} & \textbf{51.41} & 12.12 & 16.85 & 55.34 & 55.24 & 36.48 \\
\bottomrule[1.5pt]
\end{tabular}
}
\end{table}

\subsubsection{Inter-Annotator Agreement.}
\label{sec:inter_annotator}
Ego-Exo4D provides 2--5 expert annotators per take, giving a first empirical ceiling for grading quality on this benchmark.
For each pair of experts annotating the same take, we match their timestamps using the same matching radii as our evaluation protocol ($\delta \in \{0.25, 0.5, 1.0\}$\,s).
Only 13--27\% of expert annotations have an inter-expert counterpart across these radii.
On the co-located subset, averaged across radii, inter-expert balanced accuracy is 64.6 with a Cohen's $\kappa$ of 0.29 (fair agreement).
Our method's BA of 60.40 reaches 94\% of this empirical ceiling, suggesting that further gains in balanced accuracy on this benchmark are increasingly limited by label subjectivity rather than model capacity, and that mAP$_\text{S}$ remains the more discriminative metric for measuring future progress.

\subsection{Generalization to HoloAssist}
\label{sec:holoassist}
To test generalization beyond Ego-Exo4D, we evaluate \projmethod on HoloAssist~\cite{wang2023holoassist}, an egocentric mistake-detection benchmark.
While HoloAssist's official protocol performs classification on \emph{pre-segmented} clips, \projmethod extends this protocol by \emph{jointly} performing action detection and classification on \emph{untrimmed} video.
On HoloAssist's official classification metric, \projmethod reaches F1=53.3, close to the SOTA (DR-MoE~\cite{han2025dual}, F1=57.0; \cref{tab:holoassist}).
We additionally extend HoloAssist to action localization, which has not been previously evaluated on this dataset: our modules more than double ActionFormer's detection (mAP$_\text{S}$: 3.33$\to$7.24), with absolute performance consistent with the hand-manipulation scenarios (Cooking, Music) in Ego-Exo4D.

\begin{table}[t]
    \centering
    \caption{Mistake detection on HoloAssist~\cite{wang2023holoassist}. 
    \projmethod performs joint action detection and classification on \emph{untrimmed} video, while baselines classify on \emph{pre-segmented} clips. 
    PC/RC and PM/RM denote precision/recall for the Correct and Mistake classes, respectively.}
    \label{tab:holoassist}
    \vspace{-2mm}
    \begin{tabular}{@{}l ccccc@{}}
        \toprule[1.5pt]
        \textbf{Model} & F1 & PC & RC & PM & RM \\
        \midrule
        \midrule
        Random                               & 27.7 & 60.9 & 10.2 & \textbf{15.0} & 46.2 \\
        HoloAssist~\cite{wang2023holoassist}  & 36.2 & 85.5 & 43.1 & 9.7  & 11.5 \\
        DR-MoE~\cite{han2025dual}            & \textbf{57.0} & \textbf{97.0} & 60.0 & 8.0  & \textbf{63.0} \\
        SkillSpotter (ours)                   & 53.3 & 95.5 & \textbf{90.7} & 10.6 & 20.7 \\
        \bottomrule[1.5pt]
    \end{tabular}
\end{table}

\subsection{Discussion and Limitations}
\label{sec:limitations}
Performance varies substantially across scenarios: mAP$_\text{S}$ ranges from 1.57 (Music) to 40.76 (Basketball), indicating that the current method is not uniformly robust across activity types.
We interpret this variation together with strong scenario-size imbalance (3--46 videos and 191--2{,}409 annotations in our filtered evaluation split; see \cref{tab:supp_scenario_stats}, suppl.).
Accordingly, we treat Soccer (3 videos) as a low-sample regime with high metric variance and avoid strong scenario-specific conclusions, similar to the findings in egoPPG~\cite{braun2025egoppg}.
At the same time, persistent errors in high-volume scenarios such as Cooking and Music indicate that data scarcity alone does not explain the remaining failure cases.
We observe larger gains in scenarios with coarse full-body motion (e.g., Basketball and Rock Climbing) than in scenarios dominated by subtle manipulation cues (e.g., Cooking).
Cross-view attention recovers Ego+Exos BA to single-view levels (60.39) but does not surpass them, indicating that effective multi-view fusion for skill grading remains an open challenge~\cite{li2021ego}.
However, resolving the classification collapse is a necessary prerequisite for future multi-view improvements.
Our HoloAssist results confirm that \projmethod and its modules generalize beyond Ego-Exo4D, though broader cross-dataset evaluation remains an open direction.
Furthermore, our inter-annotator agreement analy{}sis yields an empirical BA ceiling of 64.6 for this benchmark, which our method's 60.40 BA approaches (94\%).
This indicates that skill assessment is inherently subjective and that mAP$_\text{S}$ remains the more discriminative metric for measuring future progress on EgoExo4D.
Promising directions for future work include incorporating physiological signals from egocentric vision, such as heart rate~\cite{braun2025egoppg, braun2024suboptimal}, heart-rate variability~\cite{demirel2026egohrv,demirelHRV}, electrodermal activity~\cite{braun2023video, braun2024sympcam}, emotional states~\cite{jammot2025egoemotion}, multi-view recordings of motions and activities~\cite{sener2022assembly101,hollidt2024egosimegocentricmulti}, and pose-object interaction features to address the remaining failure scenarios. 
End-to-end training from raw video could further jointly optimize feature extraction and temporal modeling.

\section{Conclusion}
\label{sec:conclusion}
We have introduced \projmethod, a pose-aware multi-view framework that jointly localizes and grades skilled actions in untrimmed ego-exo video through three components specifically designed for this task: adaptive temporal suppression, gated 3D body pose fusion, and bidirectional cross-view attention.
On the Ego-Exo4D demonstration proficiency benchmark, \projmethod improves class-specific mAP from 12.40 to 21.82 and balanced accuracy from 55.99\% to 60.40\% compared to the best baseline, with consistent gains across all viewpoints. 
Our adaptive suppression and pose fusion modules transfer to other backbones with similarly substantial improvements and \projmethod itself generalizes beyond Ego-Exo4D to HoloAssist.
The gains are thus not architecture- or dataset-specific.

Performance remains uneven across scenarios, with the strongest results in activities dominated by coarse full-body motion and weaker results in scenarios driven by subtle manipulation cues (Cooking and Music), indicating that the current appearance-plus-body-pose representation is insufficient for some tasks.
We additionally establish the first empirical inter-annotator agreement ceiling for this benchmark (BA: 64.6, Cohen's $\kappa$=0.29).
\projmethod reaches 94\% of this ceiling, which suggests that further BA gains are increasingly limited by label subjectivity rather than model capacity.
Incorporating pose-object interaction features and extending the framework toward natural-language feedback are promising directions for future work.


\bibliographystyle{splncs04}
\bibliography{main}

\clearpage
\appendix
\renewcommand{\theHsection}{Appendix.\thesection}


\section{Computational Cost}
\label{sec:supp_efficiency}

\cref{tab:supp_efficiency} reports model size, FLOPs, throughput, and latency for each ablation configuration.
Adaptive suppression adds negligible overhead (+0.07M parameters, $<$1 GFLOPs).
Pose fusion introduces a separate pose backbone, increasing parameters by $\sim$13M and reducing FPS from 46.80 to 36.90 (Ego).
Cross-view attention is the most expensive addition, bringing the full Ego+Exos model to 67.37M parameters and 19.70 FPS.
FPS/latency reflect model inference only.
They do not include video decoding, Omnivore feature extraction, and pose estimation.

\begin{table}
    \centering
    \caption{Computational cost per ablation configuration. Params in millions (M), latency (Lat.) in milliseconds. FPS and latency measured on a single NVIDIA H200 GPU with batch size 1.}
    \label{tab:supp_efficiency}
    \resizebox{\columnwidth}{!}{
    \begin{tabular}{@{}l cccc cccc cccc@{}}
        \toprule[1.5pt]
        & \multicolumn{4}{c}{\textbf{Ego}} & \multicolumn{4}{c}{\textbf{Exo}} & \multicolumn{4}{c}{\textbf{Ego+Exos}} \\
        \cmidrule(lr){2-5} \cmidrule(lr){6-9} \cmidrule(lr){10-13}
        \textbf{Configuration} & Params & GFLOPs & FPS & Lat. & Params & GFLOPs & FPS & Lat. & Params & GFLOPs & FPS & Lat. \\
        \midrule
        \midrule
        ActionFormer & \textbf{33.18} & \textbf{32.18} & \textbf{49.50} & \textbf{20.20} & \textbf{40.26} & \textbf{46.67} & \textbf{49.30} & \textbf{20.30} & \textbf{42.62} & \textbf{51.50} & \textbf{49.30} & \textbf{20.30} \\
        \quad + Adaptive Supp. & 33.25 & 32.72 & 46.80 & 21.40 & 40.33 & 47.21 & 46.70 & 21.40 & 42.69 & 52.04 & 46.90 & 21.30 \\
        \quad + Pose Fusion & 46.35 & 40.94 & 36.90 & 27.10 & 53.43 & 55.43 & 37.30 & 26.80 & 55.79 & 60.26 & 31.30 & 32.00 \\
        \quad + Cross-View Attn & -- & -- & -- & -- & -- & -- & -- & -- & 67.37 & 77.38 & 19.70 & 50.80 \\
        \bottomrule[1.5pt]
    \end{tabular}
    }
\end{table}

\section{TadTR Learning Rate and Query Count}
\label{sec:supp_tadtr}

In contrast to feature pyramid methods (ActionFormer, TriDet, CausalTAD, \projmethod) where every temporal position serves as a candidate detection, DETR-based detectors use a fixed set of learned queries, each predicting one event, imposing a hard upper bound on the number of detections.
Ego-Exo4D takes contain 42.3 annotations on average (median: 38, max: 266).
The default query count of TadTR~\cite{liu2022end} is 40, which covers only 54.1\% of takes.
100 queries cover 96.1\% and 200 queries 99.7\%.
\cref{tab:supp_tadtr} ablates both the query count (40, 100, 200) and the learning rate ($10^{-3}$ as used by all anchor-free models in our benchmark, and $10^{-4}$ following the standard DETR protocol).

The best detection result ($\text{lr}{=}10^{-3}$, 200 queries, Ego mAP$_\text{S}$: 7.79) remains far below ActionFormer (12.40) and our method (21.82), reported in \cref{tab:main}.
At $\text{lr}{=}10^{-4}$, performance degrades across all query counts, with 200 queries collapsing entirely (Ego mAP$_\text{A}$: 0.66).

\begin{table}
\centering
\caption{TadTR~\cite{liu2022end} ablation over two different learning rates and three different number of queries. 
Takes contain 42.3 annotations on average; 40/100/200 queries cover 54.1\%/96.1\%/99.7\% of takes.}
\label{tab:supp_tadtr}
\resizebox{\columnwidth}{!}{
\begin{tabular}{@{}cc cccc cccc cccc@{}}
\toprule[1.5pt]
& & \multicolumn{4}{c}{\textbf{Ego}} & \multicolumn{4}{c}{\textbf{Exo}} & \multicolumn{4}{c}{\textbf{Ego+Exos}} \\
\cmidrule(lr){3-6} \cmidrule(lr){7-10} \cmidrule(lr){11-14}
\textbf{lr} & \textbf{Queries} & mAP$_\text{S}$ & mAP$_\text{A}$ & BA & F1 & mAP$_\text{S}$ & mAP$_\text{A}$ & BA & F1 & mAP$_\text{S}$ & mAP$_\text{A}$ & BA & F1 \\
\midrule
\midrule
$10^{-3}$ & 40 & 5.02 & 6.59 & 49.84 & 49.60 & 6.10 & 8.61 & 50.53 & 50.51 & \textbf{6.51} & \textbf{8.89} & 51.03 & 50.80 \\
$10^{-4}$ & 40 & 3.47 & 5.60 & 52.52 & 52.34 & 1.74 & 2.91 & 49.18 & 48.47 & 5.46 & 7.79 & 51.29 & 51.28 \\
$10^{-3}$ & 100 & 6.12 & 8.57 & 52.22 & 51.74 & \textbf{7.17} & \textbf{11.09} & 50.67 & 50.38 & 3.53 & 5.33 & 49.59 & 46.95 \\
$10^{-4}$ & 100 & 2.77 & 5.14 & 53.63 & 52.60 & 4.28 & 7.41 & 52.43 & 51.51 & 2.63 & 5.58 & 51.15 & 51.11 \\
$10^{-3}$ & 200 & \textbf{7.79} & \textbf{10.79} & 49.68 & 48.03 & 6.37 & 10.48 & 52.31 & 51.25 & 4.12 & 6.74 & 52.81 & 48.23 \\
$10^{-4}$ & 200 & 0.35 & 0.66 & \textbf{55.73} & \textbf{55.64} & 2.51 & 4.97 & \textbf{55.03} & \textbf{54.92} & 1.79 & 3.89 & \textbf{54.69} & \textbf{54.67} \\
\bottomrule[1.5pt]
\end{tabular}
}
\end{table}

\section{Single-View vs.\ Concatenated Features}
\label{sec:supp_single_view}

The Ego-Exo4D baseline implementation~\cite{grauman2024ego} concatenates all available camera features into a single representation for the Exo and Ego+Exos views.
For evaluation completeness, \cref{tab:supp_views} compares this concatenation strategy against single-view processing, where each camera view is treated as an independent sample.
Concatenation improves detection mAP for most baselines but does not consistently improve classification.
For our method, concatenation yields the highest Ego+Exos mAP$_\text{S}$ (21.99) but BA drops (45.77), similar to TriDet~\cite{shi2023tridet} and ActionFormer~\cite{zhang2022actionformer}.
Our introduced cross-view attention recovers BA to 60.39 with minimal detection loss.

\begin{table}
\centering
\caption{Single-view vs.\ concatenated features for Exo and Ego+Exos settings. Ego results are identical across configurations and omitted. In the single-view setting, each exo camera is treated as an independent sample.}
\label{tab:supp_views}
\resizebox{\columnwidth}{!}{
\begin{tabular}{@{}ll cccc cccc@{}}
\toprule[1.5pt]
& & \multicolumn{4}{c}{\textbf{Exo}} & \multicolumn{4}{c}{\textbf{Ego+Exos}} \\
\cmidrule(lr){3-6} \cmidrule(lr){7-10}
\textbf{Model} & \textbf{View} & mAP$_\text{S}$ & mAP$_\text{A}$ & BA & F1 & mAP$_\text{S}$ & mAP$_\text{A}$ & BA & F1 \\
\midrule
\midrule
TadTR~\cite{liu2022end} & Single & 9.09 & 13.57 & 51.07 & 50.89 & 8.26 & 12.69 & 50.72 & 50.11 \\
TadTR~\cite{liu2022end} & Concat. & 6.37 & 10.48 & 52.31 & 34.63 & 4.12 & 6.74 & 52.81 & 48.23 \\
\midrule
TriDet~\cite{shi2023tridet} & Single & 10.42 & 13.93 & 53.33 & 52.37 & 12.38 & 16.10 & 55.32 & 55.15 \\
TriDet~\cite{shi2023tridet} & Concat. & 8.99 & 12.16 & 50.06 & 36.78 & 8.23 & 11.78 & 48.92 & 48.77 \\
\midrule
CausalTAD~\cite{liu2024harnessing} & Single & 8.08 & 10.14 & 53.05 & 52.32 & 8.15 & 9.87 & 52.04 & 51.41 \\
CausalTAD~\cite{liu2024harnessing} & Concat. & 11.82 & 14.05 & 50.78 & 50.41 & 13.16 & 17.21 & 54.98 & 54.95 \\
\midrule
ActionFormer~\cite{zhang2022actionformer} & Single & 9.49 & 11.77 & 51.31 & 50.91 & 8.15 & 10.21 & 53.33 & 53.00 \\
ActionFormer~\cite{zhang2022actionformer} & Concat. & 13.18 & 18.25 & 55.03 & 55.03 & 13.82 & 17.85 & 50.34 & 50.17 \\
\midrule
Ours & Single & 20.33 & \textbf{27.63} & 56.78 & 56.76 & 20.48 & 27.91 & 55.93 & 55.86 \\
Ours & Concat. & \textbf{21.12} & 27.47 & \textbf{60.59} & \textbf{60.55} & \textbf{21.99} & \textbf{28.04} & 45.77 & 43.57 \\
Ours & Cross-view & -- & -- & -- & -- & 21.34 & 28.01 & \textbf{60.39} & \textbf{59.37} \\
\bottomrule[1.5pt]
\end{tabular}
}
\end{table}

\section{Soft NMS Sigma Sweep}
\label{sec:supp_soft_nms_sigma_sweep}

\cref{tab:supp_sigma} reports ActionFormer~\cite{zhang2022actionformer} with Soft NMS at varying $\sigma$ values.
Detection metrics (mAP$_\text{A}$, mAP$_\text{S}$) decrease monotonically with increasing $\sigma$.
No fixed $\sigma$ value can reach the performance when not using any NMS (No NMS).
BA peaks at $\sigma{=}0.5$ (Ego: 55.34) but the margin over No NMS (54.81) is small.

\begin{table}
    \centering
    \caption{Soft NMS $\sigma$ sweep on ActionFormer. All rows use the same trained model with different post-processing.}
    \label{tab:supp_sigma}
    \resizebox{\columnwidth}{!}{
    \begin{tabular}{@{}l cccc cccc cccc@{}}
        \toprule[1.5pt]
        & \multicolumn{4}{c}{\textbf{Ego}} & \multicolumn{4}{c}{\textbf{Exo}} & \multicolumn{4}{c}{\textbf{Ego+Exos}} \\
        \cmidrule(lr){2-5} \cmidrule(lr){6-9} \cmidrule(lr){10-13}
        $\sigma$ & mAP$_\text{S}$ & mAP$_\text{A}$ & BA & F1 & mAP$_\text{S}$ & mAP$_\text{A}$ & BA & F1 & mAP$_\text{S}$ & mAP$_\text{A}$ & BA & F1 \\
        \midrule
        \midrule
        No NMS & \textbf{13.96} & \textbf{22.50} & 54.81 & 54.26 & \textbf{15.42} & \textbf{24.47} & 54.85 & 54.65 & \textbf{17.30} & \textbf{24.12} & \textbf{51.79} & \textbf{51.40} \\
        0.25 & 13.18 & 19.30 & 54.92 & 54.64 & 14.10 & 20.65 & \textbf{54.96} & \textbf{54.91} & 15.52 & 20.46 & 51.51 & 51.29 \\
        0.5 & 12.12 & 16.85 & \textbf{55.34} & \textbf{55.24} & 13.15 & 18.13 & 54.75 & 54.75 & 13.71 & 17.86 & 50.24 & 50.01 \\
        1.0 & 11.53 & 15.99 & 52.84 & 52.81 & 12.35 & 17.47 & 53.22 & 53.18 & 11.72 & 15.48 & 48.48 & 48.12 \\
        2.0 & 7.40 & 11.00 & 52.01 & 51.97 & 8.22 & 12.48 & 51.57 & 51.50 & 8.57 & 12.03 & 49.69 & 49.33 \\
        3.0 & 6.64 & 10.37 & 52.15 & 52.11 & 7.46 & 11.60 & 51.28 & 51.28 & 7.32 & 11.41 & 49.79 & 49.53 \\
        \bottomrule[1.5pt]
    \end{tabular}
    }
\end{table}

\section{Cross-Backbone Evaluation}
\label{sec:suppl_cross_backbone2}
We apply adaptive temporal suppression and gated pose fusion to all evaluated architectures except TadTR~\cite{liu2022end} to evaluate if our modules are specific to the ActionFormer backbone (see \cref{tab:cross_backbone}).
We exclude TadTR because its DETR-style set prediction removes the suppression stage entirely, making adaptive temporal suppression inapplicable.
All models use the same pre-extracted Omnivore features, training protocol, and evaluation settings as in the main paper.

\subsubsection{Adaptive Suppression Consistently Improves Detection.}
Across all backbones and all view settings, adaptive suppression substantially improves mAP$_\text{S}$ over both the Soft NMS and No NMS baselines.
These gains are consistent with those observed on ActionFormer (see \cref{tab:ablation}, main paper).

\subsubsection{Pose Fusion's Effect on Grading Is Backbone- and View-Dependent.}
Unlike adaptive suppression, gated pose fusion does not uniformly improve balanced accuracy (BA) across all backbones.
On CausalTAD and TriDet, pose fusion improves BA in every view setting (e.g., CausalTAD Ego: 54.01$\to$59.34, +5.33; TriDet Exos: 57.54$\to$60.56, +3.02).
On the remaining three backbones, pose fusion slightly degrades single-view grading: VideoMambaSuite (Ego: 56.89$\to$55.01; Exos: 57.59$\to$54.46) and TemporalMaxer (Ego: 59.95$\to$58.66; Exos: 61.27$\to$59.66) both lose BA when pose is added on Ego and Exos, while gaining mAP$_\text{S}$/mAP$_\text{A}$.
In Ego+Exos settings, however, pose fusion consistently recovers or improves grading even where it hurts single-view BA (e.g., TemporalMaxer Ego+Exos: 46.86$\to$59.42).

\subsubsection{Cross-View Attention Is Architecture-Dependent.}
Unlike ActionFormer, which suffers severe BA collapse under naive Ego+Exos concatenation (see \cref{tab:ablation}, main paper), all of the evaluated backbones achieve reasonable Ego+Exos BA with adaptive suppression and pose fusion alone.
Adding cross-view attention does not improve and can degrade performance in these cases.
We attribute this to the fact that these architectures do not exhibit the classification collapse under multi-view concatenation that we observe with ActionFormer (see \cref{tab:ablation}, Ego+Exos BA: 45.77 vs.\ 57.85 and 60.01).
Since cross-view attention specifically addresses this collapse, we apply it only to ActionFormer, where the failure mode is present.

\begin{table}
    \centering
    \caption{Component ablation on the VideoMambaSuite~\cite{chen2026video} backbone.
    Each row progressively adds one module.
    $^*$~Uses cross-view attention for Ego+Exos.}
    \label{tab:ablation_videomambasuite}
    \resizebox{\columnwidth}{!}{
    \begin{tabular}{@{}l cccc cccc cccc@{}}
        \toprule[1.5pt]
        & \multicolumn{4}{c}{\textbf{Ego}} & \multicolumn{4}{c}{\textbf{Exos}} & \multicolumn{4}{c}{\textbf{Ego+Exos}} \\
        \cmidrule(lr){2-5} \cmidrule(lr){6-9} \cmidrule(lr){10-13}
        \textbf{Configuration} & mAP$_\text{S}$ & mAP$_\text{A}$ & BA & F1 & mAP$_\text{S}$ & mAP$_\text{A}$ & BA & F1 & mAP$_\text{S}$ & mAP$_\text{A}$ & BA & F1 \\
        \midrule
        \midrule
        VideoMambaSuite (Soft NMS) & 7.63 & 8.65 & 49.51 & 49.35 & 7.06 & 8.17 & 46.03 & 43.88 & 3.69 & 3.91 & 52.70 & 51.96 \\
        VideoMambaSuite (No NMS)   & 13.60 & 16.22 & 53.55 & 53.23 & 11.44 & 15.70 & 47.79 & 40.48 & 6.91 & 9.22 & 52.02 & 47.96 \\
        \midrule
        + Adaptive Supp.           & 17.78 & 23.38 & \textbf{56.89} & \textbf{56.87} & \textbf{20.08} & \textbf{26.62} & \textbf{57.59} & \textbf{57.57} & 20.67 & 26.72 & 56.77 & 56.05 \\
        + Pose                     & \textbf{21.39} & \textbf{27.11} & 55.01 & 53.50 & 18.66 & 26.33 & 54.46 & 54.06 & 19.91 & 26.03 & \textbf{56.92} & \textbf{56.90} \\
        + Cross-View Attn$^*$      & -- & -- & -- & -- & -- & -- & -- & -- & \textbf{21.51} & \textbf{28.48} & 55.52 & 55.05 \\
        \bottomrule[1.5pt]
    \end{tabular}
    }
\end{table}

\begin{table}
    \centering
    \caption{Component ablation on the TriDet~\cite{shi2023tridet} backbone.
    Each row progressively adds one module.
    $^*$~Uses cross-view attention for Ego+Exos.}
    \label{tab:ablation_tridet}
    \resizebox{\columnwidth}{!}{
    \begin{tabular}{@{}l cccc cccc cccc@{}}
        \toprule[1.5pt]
        & \multicolumn{4}{c}{\textbf{Ego}} & \multicolumn{4}{c}{\textbf{Exos}} & \multicolumn{4}{c}{\textbf{Ego+Exos}} \\
        \cmidrule(lr){2-5} \cmidrule(lr){6-9} \cmidrule(lr){10-13}
        \textbf{Configuration} & mAP$_\text{S}$ & mAP$_\text{A}$ & BA & F1 & mAP$_\text{S}$ & mAP$_\text{A}$ & BA & F1 & mAP$_\text{S}$ & mAP$_\text{A}$ & BA & F1 \\
        \midrule
        \midrule
        TriDet (Soft NMS) & 10.35 & 14.79 & 49.17 & 40.10 & 8.99 & 12.16 & 50.06 & 36.78 & 8.23 & 11.78 & 48.92 & 48.77 \\
        TriDet (No NMS) & 11.80 & 17.80 & 52.03 & 50.56 & 11.47 & 15.45 & 48.53 & 48.49 & 9.46 & 14.71 & 48.83 & 48.25 \\
        \midrule
        + Adaptive Supp. & 18.84 & 25.98 & 58.89 & 57.30 & 19.03 & 23.85 & 57.54 & 57.15 & \textbf{19.74} & 23.94 & 53.37 & 53.10 \\
        + Pose & \textbf{20.47} & \textbf{26.91} & \textbf{59.98} & \textbf{58.77} & \textbf{19.21} & \textbf{24.32} & \textbf{60.56} & \textbf{59.93} & 19.33 & \textbf{25.56} & \textbf{57.85} & \textbf{57.84} \\
        + Cross-View Attn$^*$ & -- & -- & -- & -- & -- & -- & -- & -- & 18.70 & 23.63 & 54.08 & 53.61 \\
        \bottomrule[1.5pt]
    \end{tabular}
    }
\end{table}

\begin{table}
    \centering
    \caption{Component ablation on the TemporalMaxer~\cite{tang2023temporalmaxer} backbone.
    Each row progressively adds one module.
    $^*$~Uses cross-view attention for Ego+Exos.}
    \label{tab:ablation_temporalmaxer}
    \resizebox{\columnwidth}{!}{
    \begin{tabular}{@{}l cccc cccc cccc@{}}
        \toprule[1.5pt]
        & \multicolumn{4}{c}{\textbf{Ego}} & \multicolumn{4}{c}{\textbf{Exos}} & \multicolumn{4}{c}{\textbf{Ego+Exos}} \\
        \cmidrule(lr){2-5} \cmidrule(lr){6-9} \cmidrule(lr){10-13}
        \textbf{Configuration} & mAP$_\text{S}$ & mAP$_\text{A}$ & BA & F1 & mAP$_\text{S}$ & mAP$_\text{A}$ & BA & F1 & mAP$_\text{S}$ & mAP$_\text{A}$ & BA & F1 \\
        \midrule
        \midrule
        TemporalMaxer (Soft NMS) & 12.34 & 16.69 & 54.34 & 53.90 & 10.38 & 15.17 & 52.18 & 52.16 & 11.27 & 15.75 & 50.09 & 49.58 \\
        TemporalMaxer (No NMS)   & 14.76 & 22.83 & 53.89 & 53.12 & 11.56 & 20.53 & 52.57 & 52.55 & 13.20 & 22.42 & 52.07 & 50.46 \\
        \midrule
        + Adaptive Supp.         & \textbf{20.51} & 25.90 & \textbf{59.95} & \textbf{59.65} & \textbf{21.27} & 25.82 & \textbf{61.27} & \textbf{61.27} & 19.56 & 26.36 & 46.86 & 44.73 \\
        + Pose                   & 20.15 & \textbf{27.85} & 58.66 & 57.48 & 20.65 & \textbf{27.01} & 59.66 & 59.66 & \textbf{20.55} & \textbf{27.38} & \textbf{59.42} & \textbf{59.34} \\
        + Cross-View Attn$^*$    & -- & -- & -- & -- & -- & -- & -- & -- & 19.61 & 27.46 & 58.38 & 56.57 \\
        \bottomrule[1.5pt]
    \end{tabular}
    }
\end{table}

\begin{table}
    \centering
    \caption{Component ablation on the CausalTAD~\cite{liu2024harnessing} backbone.
    Each row progressively adds one module.
    $^*$~Uses cross-view attention for Ego+Exos.}
    \label{tab:ablation_causaltad}
    \resizebox{\columnwidth}{!}{
    \begin{tabular}{@{}l cccc cccc cccc@{}}
        \toprule[1.5pt]
        & \multicolumn{4}{c}{\textbf{Ego}} & \multicolumn{4}{c}{\textbf{Exos}} & \multicolumn{4}{c}{\textbf{Ego+Exos}} \\
        \cmidrule(lr){2-5} \cmidrule(lr){6-9} \cmidrule(lr){10-13}
        \textbf{Configuration} & mAP$_\text{S}$ & mAP$_\text{A}$ & BA & F1 & mAP$_\text{S}$ & mAP$_\text{A}$ & BA & F1 & mAP$_\text{S}$ & mAP$_\text{A}$ & BA & F1 \\
        \midrule
        \midrule
        CausalTAD (Soft NMS) & 11.42 & 16.07 & 52.86 & 52.07 & 11.82 & 14.05 & 50.78 & 50.41 & 13.16 & 17.21 & 54.98 & 54.95 \\
        CausalTAD (No NMS) & 12.77 & 22.71 & 51.96 & 50.20 & 12.57 & 18.91 & 53.16 & 52.87 & 12.69 & 21.45 & 54.63 & 54.47 \\
        \midrule
        + Adaptive Supp. & 19.64 & 25.78 & 54.01 & 52.79 & 17.68 & 22.55 & 55.67 & 54.96 & 18.90 & 26.42 & 58.15 & 55.85 \\
        + Pose & \textbf{21.36} & \textbf{27.46} & \textbf{59.34} & \textbf{59.23} & \textbf{20.76} & \textbf{25.81} & \textbf{60.53} & \textbf{60.36} & 20.02 & 26.83 & \textbf{60.01} & \textbf{59.51} \\
        + Cross-View Attn$^*$ & -- & -- & -- & -- & -- & -- & -- & -- & \textbf{23.29} & \textbf{28.41} & 59.11 & 59.45 \\
        \bottomrule[1.5pt]
    \end{tabular}
    }
\end{table}

\begin{table}
    \centering
    \caption{Component ablation on the DyFADet~\cite{yang2024dyfadet} backbone.
    Each row progressively adds one module.
    $^*$~Uses cross-view attention for Ego+Exos.}
    \label{tab:ablation_dyfadet}
    \resizebox{\columnwidth}{!}{
    \begin{tabular}{@{}l cccc cccc cccc@{}}
        \toprule[1.5pt]
        & \multicolumn{4}{c}{\textbf{Ego}} & \multicolumn{4}{c}{\textbf{Exos}} & \multicolumn{4}{c}{\textbf{Ego+Exos}} \\
        \cmidrule(lr){2-5} \cmidrule(lr){6-9} \cmidrule(lr){10-13}
        \textbf{Configuration} & mAP$_\text{S}$ & mAP$_\text{A}$ & BA & F1 & mAP$_\text{S}$ & mAP$_\text{A}$ & BA & F1 & mAP$_\text{S}$ & mAP$_\text{A}$ & BA & F1 \\
        \midrule
        \midrule
        DyFADet (Soft NMS) & 10.09 & 12.53 & 48.89 & 46.75 & 3.57 & 4.48 & 49.52 & 47.72 & 3.18 & 5.64 & 47.63 & 35.45 \\
        DyFADet (No NMS)   & 12.74 & 16.26 & 50.14 & 46.48 & 3.68 & 4.69 & 47.23 & 45.00 & 3.69 & 5.57 & 47.54 & 40.41 \\
        \midrule
        + Adaptive Supp.   & 21.14 & 26.15 & 55.33 & 54.75 & 16.08 & 21.38 & \textbf{54.36} & \textbf{52.17} & 18.79 & 23.41 & 55.94 & 55.48 \\
        + Pose             & \textbf{21.56} & \textbf{28.68} & \textbf{56.77} & \textbf{56.51} & \textbf{20.16} & \textbf{25.22} & 41.83 & 36.71 & 20.94 & 27.25 & \textbf{58.86} & \textbf{58.49} \\
        + Cross-View Attn$^*$& -- & -- & -- & -- & -- & -- & -- & -- & \textbf{23.18} & \textbf{28.62} & 57.50 & 57.46 \\
        \bottomrule[1.5pt]
    \end{tabular}
    }
\end{table}

\section{AQA Drop-In Classification Heads}
\label{sec:suppl_aqa_heads}
Action quality assessment (AQA) methods regress a scalar quality score from pre-segmented clips but provide no localization.
To test whether AQA heads perform better on skill action grading than our classification head, we replace \projmethod's classification head with three AQA heads, adapting only the components required by our binary good/tip classification task: TechCoach~\cite{li2026techcoach}, GDLT~\cite{xu2022likert}, and USDL~\cite{tang2020uncertainty}.
Our default head wins on grading quality (BA, F1) in all three view settings by a wide margin, while detection quality is slightly better using USDL in most settings.
TechCoach's performance on grading is closest to ours on Ego (BA: 58.87), but is unstable across views, collapsing on Exos (BA: 43.11).

\begin{table}
    \centering
    \caption{\projmethod with drop-in AQA classification heads across all three view settings. Our default head achieves the best grading quality (BA, F1) in every setting; USDL achieves the best detection (mAP$_\text{S}$, mAP$_\text{A}$) in most settings despite trailing substantially in grading.}
    \label{tab:aqa_heads}
    \resizebox{\columnwidth}{!}{
    \begin{tabular}{@{}l cccc cccc cccc@{}}
        \toprule[1.5pt]
        & \multicolumn{4}{c}{\textbf{Ego}} & \multicolumn{4}{c}{\textbf{Exos}} & \multicolumn{4}{c}{\textbf{Ego+Exos}} \\
        \cmidrule(lr){2-5} \cmidrule(lr){6-9} \cmidrule(lr){10-13}
        \textbf{Head Type} & mAP$_\text{S}$ & mAP$_\text{A}$ & BA & F1 & mAP$_\text{S}$ & mAP$_\text{A}$ & BA & F1 & mAP$_\text{S}$ & mAP$_\text{A}$ & BA & F1 \\
        \midrule
        \midrule
        TechCoach~\cite{li2026techcoach} & 20.10 & 27.15 & 58.87 & 57.60 & 19.66 & 27.72 & 43.11 & 37.77 & 18.84 & 26.16 & 58.90 & 58.38 \\
        GDLT~\cite{xu2022likert}           & 21.52 & 27.28 & 42.48 & 38.53 & 19.37 & 26.33 & 46.77 & 46.17 & 20.82 & 26.28 & 40.68 & 36.10 \\
        USDL~\cite{tang2020uncertainty}  & 21.44 & \textbf{29.12} & 49.99 & 49.14 & \textbf{21.63} & \textbf{28.47} & 50.87 & 50.03 & \textbf{21.67} & \textbf{28.11} & 54.78 & 54.60 \\
        Ours (default)                   & \textbf{21.82} & 27.89 & \textbf{60.40} & \textbf{60.02} & 21.12 & 27.47 & \textbf{60.59} & \textbf{60.55} & 21.34 & 28.01 & \textbf{60.39} & \textbf{59.37} \\
        \bottomrule[1.5pt]
    \end{tabular}
    }
\end{table}

\section{Predicted vs.\ Ground Truth Pose}
\label{sec:supp_pose}
\cref{tab:supp_pose} compares three pose configurations: training and testing on predicted pose, training on ground truth with predicted pose at test time (our default), and training and testing on ground truth pose (oracle).
Training on ground truth pose outperforms training on predicted pose, even when both test on predicted input (e.g., Ego mAP$_\text{S}$: 21.82 vs.\ 19.96), indicating that the model learns cleaner biomechanical representations from ground truth annotations.
The gap between our default setting and the oracle is small across all configurations (e.g., Ego BA: 60.40 vs.\ 61.39, $\Delta{=}0.99$), confirming that predicted pose quality is sufficient for effective skill grading during inference.

\begin{table}
\centering
\caption{Impact of pose quality during training and inference.
Predicted ego pose via the official Ego-Exo4D baseline~\cite{grauman2024ego} (PA-MPJPE: 10.70\,cm); predicted exo pose via ViTPose~\cite{xu2022vitpose} with multi-view triangulation (PA-MPJPE: 17.98\,cm).
We report Procrustes-aligned MPJPE as our pose features (joint angles, distances, velocities) are invariant to the global coordinate frame.}
\label{tab:supp_pose}
\resizebox{\columnwidth}{!}{
\begin{tabular}{@{}ll cccc cccc cccc@{}}
\toprule[1.5pt]
& & \multicolumn{4}{c}{\textbf{Ego}} & \multicolumn{4}{c}{\textbf{Exo}} & \multicolumn{4}{c}{\textbf{Ego+Exos}} \\
\cmidrule(lr){3-6} \cmidrule(lr){7-10} \cmidrule(lr){11-14}
\textbf{Training} & \textbf{Testing} & mAP$_\text{S}$ & mAP$_\text{A}$ & BA & F1 & mAP$_\text{S}$ & mAP$_\text{A}$ & BA & F1 & mAP$_\text{S}$ & mAP$_\text{A}$ & BA & F1 \\
\midrule
\midrule
Predicted & Predicted & 19.96 & 27.26 & 60.31 & 60.01 & 20.36 & 27.13 & 59.61 & 58.92 & 20.74 & 27.69 & 58.71 & 57.77 \\
Ground truth & Predicted & 21.82 & 27.89 & 60.40 & 60.02 & 21.12 & 27.47 & 60.59 & 60.55 & 21.34 & 28.01 & 60.39 & 59.37 \\
Ground truth & Ground truth & \textbf{22.09} & \textbf{28.00} & \textbf{61.39} & \textbf{61.04} & \textbf{21.74} & \textbf{27.70} & \textbf{60.87} & \textbf{60.82} & \textbf{21.60} & \textbf{28.03} & \textbf{60.59} & \textbf{59.72} \\
\bottomrule[1.5pt]
\end{tabular}
}
\end{table}



\section{Per-Scenario Radius Distributions}
\label{sec:supp_radius_boxplot}
\begin{figure}[t]
  \centering
  \includegraphics[width=\linewidth]{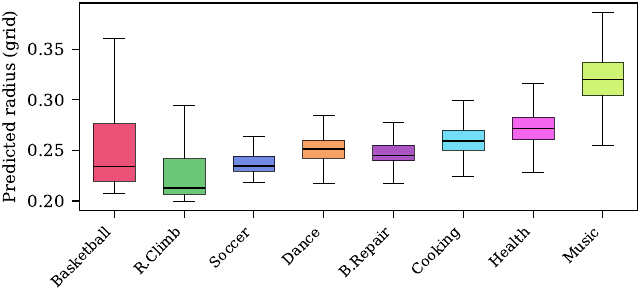}
  \caption{%
    Distribution of learned suppression radii per scenario.
    Scenarios with denser event distributions (Basketball, Rock Climbing) produce smaller radii, while sparse scenarios (Music, Health) produce larger radii.
    Box plots show median (orange line), interquartile range (box), and 1.5$\times$IQR whiskers.
  }
  \label{fig:supp_radius_boxplot}
\end{figure}

\cref{fig:supp_radius_boxplot} shows the distribution of learned suppression radii across all eight scenarios.
The suppression head learns scenario-specific radii without explicit supervision on the radius values---only the auxiliary suppression loss (\cref{sec:suppression}) provides a training signal.
Scenarios with sparse, well-separated events (Music, Health) produce larger median radii (0.32 and 0.27 grid units, respectively), allowing broader suppression of isolated false positives.
Conversely, dense-event scenarios (Rock Climbing, Basketball) yield smaller median radii (0.21 and 0.23) but exhibit wider spread, reflecting the need for context-dependent radii: detections near temporal clusters require tight suppression to preserve neighbors, while isolated detections permit larger radii.
This adaptive behavior explains the failure mode of fixed-radius NMS: a single radius cannot simultaneously avoid over-suppression in dense scenarios and under-suppression in sparse ones (see \cref{fig:density_cdf}).

\section{Dataset Statistics}
\label{sec:supp_dataset_stats}

We report annotation statistics at two levels: the official Ego-Exo4D proficiency annotation files and the filtered evaluation database that our models are actually trained and evaluated on.

\subsubsection{Official Annotations.}
\cref{tab:supp_scenario_stats_official} shows the per-scenario annotation counts directly from the official annotation files released by the Ego-Exo4D benchmark.
The training split contains 556 takes with 11{,}942 annotations, and the test split contains 177 takes with 3{,}707 annotations.
Good and bad labels are near-perfectly balanced across all scenarios (approximately 50/50).

\subsubsection{Evaluation Database.}
Our preprocessing pipeline converts the official annotations into a unified database while applying several necessary filters:
(1)~takes whose take\_uid is absent from the Ego-Exo4D metadata are dropped;
(2)~takes with anomalous frame rates (outside $[10, 100]$~fps) are excluded;
and (3)~views for which the pre-extracted Omnivore feature files are missing are skipped.
These filters reduce the training set from 556 to 492 takes and the test set from 177 to 176 takes (\cref{tab:supp_scenario_stats}).
The small reduction confirms that the vast majority of officially annotated takes have complete metadata, valid frame rates, and available features.
The dropped takes are predominantly missing pre-extracted Omnivore features for one or more views.

\begin{table}[t]
  \centering
  \caption{%
    Per-scenario annotation statistics from the official Ego-Exo4D proficiency annotation files.
    Good/bad labels are near-perfectly balanced overall, with minor per-scenario deviations.
  }
  \label{tab:supp_scenario_stats_official}
  \setlength{\tabcolsep}{4pt}
  \resizebox{\columnwidth}{!}{
  \begin{tabular}{l rrrr rrrr}
    \toprule[1.5pt]
    & \multicolumn{4}{c}{\textbf{Train}} & \multicolumn{4}{c}{\textbf{Test}} \\
    \cmidrule(lr){2-5} \cmidrule(lr){6-9}
    \textbf{Scenario} & Total & Good & Bad & Takes & Total & Good & Bad & Takes \\
    \midrule
    \midrule
    Basketball       & 7{,}984 & 4{,}083 & 3{,}901 & 146 & 2{,}492 & 1{,}293 & 1{,}199 & 47 \\
    Bike Repair      & 1{,}448 & 704 & 744 & 41 & 191 & 107 & 84 & 9 \\
    Cooking          & 5{,}212 & 2{,}689 & 2{,}523 & 80 & 1{,}109 & 602 & 507 & 24 \\
    Dance            & 1{,}942 & 957 & 985 & 80 & 843 & 405 & 438 & 35 \\
    Health           & 1{,}804 & 764 & 1{,}040 & 42 & 547 & 237 & 310 & 12 \\
    Music            & 2{,}516 & 1{,}034 & 1{,}482 & 94 & 703 & 281 & 422 & 32 \\
    Rock Climbing    & 2{,}702 & 1{,}364 & 1{,}338 & 65 & 732 & 395 & 337 & 15 \\
    Soccer           & 685 & 261 & 424 & 8 & 380 & 137 & 243 & 3 \\
    \midrule
    All              & 24{,}293 & 11{,}856 & 12{,}437 & 556 & 6{,}997 & 3{,}457 & 3{,}540 & 177 \\
    \bottomrule[1.5pt]
  \end{tabular}}
\end{table}

\begin{table}[t]
  \centering
  \caption{%
    Per-scenario annotation statistics as used in our evaluation pipeline.
    Takes without all required views or pre-extracted features are excluded from the official counts (\cref{tab:supp_scenario_stats_official}).
  }
  \label{tab:supp_scenario_stats}
  \setlength{\tabcolsep}{4pt}
  \resizebox{\columnwidth}{!}{
  \begin{tabular}{l rrrr rrrr}
    \toprule[1.5pt]
    & \multicolumn{4}{c}{\textbf{Train}} & \multicolumn{4}{c}{\textbf{Test}} \\
    \cmidrule(lr){2-5} \cmidrule(lr){6-9}
    \textbf{Scenario} & Total & Good & Bad & Takes & Total & Good & Bad & Takes \\
    \midrule
    \midrule
    Basketball       & 6{,}899 & 3{,}577 & 3{,}322 & 129 & 2{,}409 & 1{,}260 & 1{,}149 & 46 \\
    Bike Repair      & 1{,}311 & 643 & 668 & 34 & 191 & 107 & 84 & 9 \\
    Cooking          & 4{,}605 & 2{,}368 & 2{,}237 & 69 & 1{,}109 & 602 & 507 & 24 \\
    Dance            & 1{,}813 & 892 & 921 & 75 & 843 & 405 & 438 & 35 \\
    Health           & 1{,}473 & 633 & 840 & 35 & 547 & 237 & 310 & 12 \\
    Music            & 2{,}080 & 826 & 1{,}254 & 83 & 703 & 281 & 422 & 32 \\
    Rock Climbing    & 2{,}445 & 1{,}217 & 1{,}228 & 61 & 732 & 395 & 337 & 15 \\
    Soccer           & 527 & 221 & 306 & 6 & 380 & 137 & 243 & 3 \\
    \midrule
    All              & 21{,}153 & 10{,}377 & 10{,}776 & 492 & 6{,}914 & 3{,}424 & 3{,}490 & 176 \\
    \bottomrule[1.5pt]
  \end{tabular}}
\end{table}

\end{document}